%% file: main.tex
\documentclass[10pt,twocolumn,letterpaper]{article}
\usepackage[letterpaper, margin=0.6in,tmargin=1.1in,bmargin=1.0in]{geometry}

\input{preamble}

\usepackage[pagebackref,breaklinks,colorlinks]{hyperref}

\usepackage{sectsty}

\sectionfont{\fontsize{12}{15}\selectfont}
\usepackage{graphicx}
\usepackage{amsmath}
\usepackage{amssymb}
\usepackage{amsfonts,amsthm}
\usepackage{booktabs}
\usepackage{subcaption}
\usepackage{xcolor}
\usepackage{tablefootnote}
\usepackage{colortbl}
\usepackage{enumitem}
\usepackage[capitalize]{cleveref}
\crefname{section}{Sec.}{Secs.}
\Crefname{section}{Section}{Sections}
\Crefname{table}{Table}{Tables}
\crefname{table}{Tab.}{Tabs.}

\newcommand{\algname}{\mbox{CTT}}
\newcommand{\fullalgname}{\mbox{Categorical Traffic Transformer}}

\definecolor{cello}{HTML}{ffe6cc}
\newcommand{\colorcello}{\cellcolor{cello}}
\newcommand{\aux}{\mathrm{aux}}
\newcommand{\lm}{\mathtt{L}}
\newcommand{\homo}{\mathtt{H}}
\newcommand{\relu}[1]{\text{ReLU}\left(#1\right)}

\newcommand\marklessfootnote[1]{
    \addtocounter{footnote}{1} %no call is made to \footnotemark.
    \footnotetext{#1}
}

\newtheorem{rem}{Remark}
\author{Authors}

\begin{document}
\interfootnotelinepenalty=10000
\title{
% \algname{}: Scene-Consistent Trajectory Prediction for Downstream Planning
% \fullalgname{}: elevating traffic modeling to better accuracy, diversity, and interpretability with tokenized latent}
% \fullalgname{} for Scene-Consistent Trajectory Prediction
% \fullalgname{} for Controllable Multimodal Trajectory Prediction
\fullalgname{}: Interpretable and Diverse Behavior Prediction with Tokenized Latent
}

\author{Yuxiao Chen$^1$ \hspace{1cm} Sander Tonkens$^{1,2}$ \hspace{1cm} Marco Pavone$^{1,3}$\\
$^1$NVIDIA Research \hspace{1cm} $^2$University of California, San Diego \hspace{1cm} $^3$Stanford University\\
{\tt\small \{yuxiaoc, stonkens, mpavone\}@nvidia.com, stonkens@ucsd.edu, pavone@stanford.edu}
% For a paper whose authors are all at the same institution,
% omit the following lines up until the closing ``}''.
% Additional authors and addresses can be added with ``\and'',
% just like the second author.
% To save space, use either the email address or home page, not both
% \and
% Boris Ivanovic\\
% NVIDIA Research\\
% {\tt\small bivanovic@nvidia.com}
% \and 
% Marco Pavone\\
% NVIDIA Research\\
% {\tt\small mpavone@nvidia.com}
\date{}
}
\maketitle

\begin{abstract}\marklessfootnote{We will release code for \algname{} once the paper review process is done.}
Adept traffic models are critical to both planning and closed-loop simulation for autonomous vehicles (AV), and key design objectives include accuracy, diverse multimodal behaviors, interpretability, and downstream compatibility. Recently, with the advent of large language models (LLMs), an additional desirable feature for traffic models is LLM compatibility. We present \fullalgname{} (\algname{}), a traffic model that outputs both continuous trajectory predictions and tokenized categorical predictions (lane modes, homotopies, etc.). The most outstanding feature of \algname{} is its fully interpretable latent space, which enables direct supervision of the latent variable from the ground truth during training and avoids mode collapse completely. As a result, \algname{} can generate diverse behaviors conditioned on different latent modes with semantic meanings while beating SOTA on prediction accuracy. In addition, \algname{}'s ability to input and output tokens enables integration with LLMs for common-sense reasoning and zero-shot generalization.
\end{abstract}
\vspace{-0.5cm}
\input{introduction}

\input{related_works}
\input{method}
\input{result}
\vspace{-0.2cm}
\input{conclusion}

{\small
\bibliographystyle{ieee_fullname.bst}
\bibliography{bibliography}
}

% \clearpage
\newpage
\include{appendix}
\end{document}

%% file: preamble.tex
\usepackage[dvipsnames]{xcolor}

%% file: introduction.tex
\section{Introduction}\label{sec:intro}

Traffic models are widely used in AV planning and closed-loop simulation and have been studied extensively. State-of-the-art (SOTA) models are predominantly learning-based \cite{salzmannIvanovicEtAl2020,GuSunEtAl2021,ChenIvanovicEtAl2022,luoParkEtAl2023}. In order to model multimodal human behaviors, the majority of traffic models generate multimodal predictions using an encoder-decoder structure with a noninterpretable latent space. Common examples include a Gaussian distribution \cite{YuanWengEtAl2021,luoParkEtAl2023,RhinehartMcAllisterEtAl2019} or a discrete distribution \cite{ChenIvanovicEtAl2022}. However, since the latent space is not interpretable, no direct supervision of the latent mode is available during training. In an end-to-end training setup, the latent distribution is learned as a byproduct of directly minimizing the multimodal prediction loss; with techniques such as the reparameterization trick \cite{KingmaWelling2013} and Gumbel softmax \cite{JangGuEtAl2017} for differentiability. Since all modes are supervised with the same ground truth (GT) trajectory, they tend to converge towards the most likely mode, a phenomenon referred to as mode collapse. Several workarounds have been proposed to mitigate mode collapse, such as Dlow \cite{yuanKitani2020} and the CVaR trick \cite{ChenIvanovicEtAl2022}, yet their efficacy is limited. For a downstream planner, it is desirable that the multimodal predictions are an accurate description of the stochastic distribution of the joint behavior of all agents involved and are distinct from each other, i.e., a small variation of motion should not constitute a new mode. In practice, traffic models trained to minimize the prediction error often fail to produce diverse multimodal predictions that cover the possible modes of scene rollout, making it challenging for a downstream planner to plan with such multimodal predictions.

In addition to a high-quality multimodal prediction, there are other important aspects of downstream compatibility. For example, it is shown in \cite{ChenKarkusEtAl2023} that a Gaussian latent space can lead to poor temporal consistency due to the need to sample from the latent space at every time step, hurting closed-loop planning performance. Using a discrete latent space can partially mitigate the problem, yet using a fixed number of discrete modes makes it difficult for the model to adapt to diverse scenarios and can lead to poor expressiveness. When traffic models are used as a simulation agent, it is desirable if the multimodal behaviors are interpretable and can be linked to semantics that humans understand.

On a different note, the world is witnessing significant progress in the field of language models, particularly large language models (LLMs) such as the GPT series \cite{brownMannEtAl2020,BubeckChandrasekaranEtAl2023} and Llama \cite{TouvronLavrilEtAl2023}, which encourages related fields such as robotics and AV to embrace the idea of embodied AI when building their next-generation autonomy stack. While LLMs demonstrate amazing capabilities in logical reasoning and common-sense knowledge, they lack domain-specific skills such as fine manipulation skills when applied to grasping and require an ``expert" policy to fill this gap \cite{DriessXiaEtAl2023palm}. In the field of AV, it is thus desirable to combine LLMs with traffic models, with the latter providing the domain knowledge of driving lacking in the former. Since LLMs operate in a tokenized world, there needs to be a communication channel between the multimodal predictions from traffic models and the semantic concepts that LLMs consume/generate, which is missing in SOTA traffic models.
\begin{figure*}
\centering
    \includegraphics[width=1\textwidth]{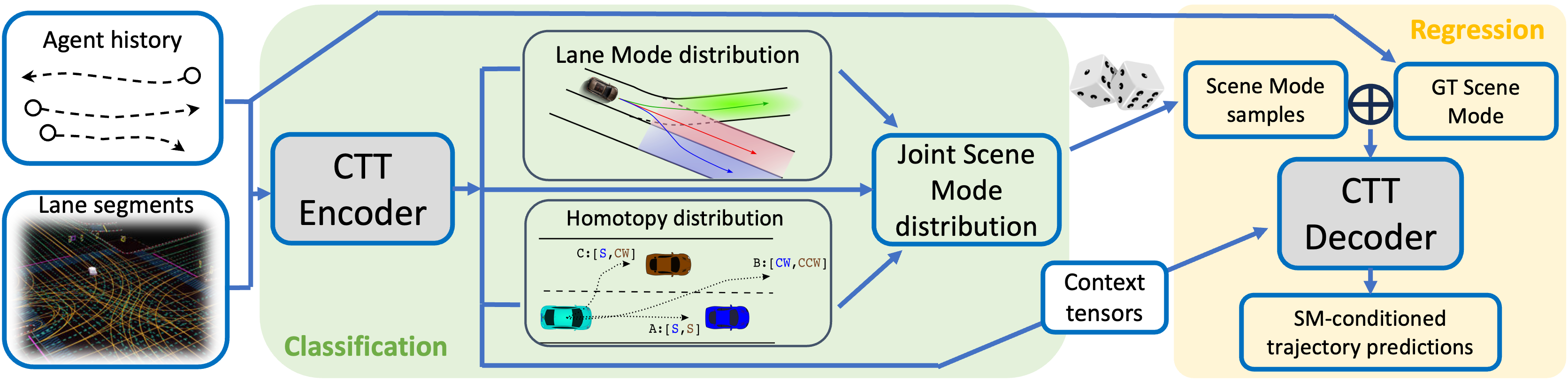}
    \caption{Model architecture of \algname{}, where the encoder predicts the Scene Mode (SM) consisting of agent2lane (a2l) and agent2agent (a2a) modes, and the decoder generates trajectory predictions conditioned on the SM samples. The ground truth SM is identified directly from the driving log and thus decouples the encoder and decoder training (except for the shared context tensors).}
\end{figure*}

Given the aforementioned challenges and opportunities, we propose a traffic model with a tokenized latent space that can be supervised directly during training. Specifically, we introduce \fullalgname{} (\algname{}), a transformer-based traffic model whose most important feature is an interpretable latent space consisting of a set of modes containing agent-to-agent interaction (a2a) modes and agent-to-lane (a2l) modes. We explicitly categorize the scene's future rollout using a product of a2a and a2l modes during training. As a result, the encoder and decoder training objectives are decoupled and the latent variable is directly supervised. Since the ground truth (GT) latent mode is directly identified from the driving log, only the trajectory prediction under the GT latent mode is supervised by the GT future trajectories. Trajectory predictions conditioned on non-GT latent modes are instead supervised by consistency losses to ensure that the decoded trajectories conform with their latent modes, which further improves mode diversity. Architecture-wise, we propose a flexible transformer architecture that is equivariant (consistency under coordinate change) thanks to Custom Edge Embedding (CEE). The transformer model is combined with Graph Neural Networks (GNNs) to enhance its capability of handling tokenized edges. The main novelties of \algname{} are:
\begin{itemize}[noitemsep,topsep=0pt]
    \item An interpretable set of modes that allows for direct supervision of the latent distribution, enabling the encoder and decoder training objectives to be decoupled.
    \item The use of consistency loss to facilitate the decoder training and improve prediction diversity.
    \item A flexible transformer and GNN architecture that is equivariant, accepts a wide range of tokenized inputs, and is compatible with language models.
\end{itemize}
We tested \algname{} on three public datasets: nuScenes \cite{CaesarBankitiEtAl2019}, nuPlan \cite{caesarKabzanEtAl2021}, and Waymo Open Dataset (WOMD) \cite{waymo_open_dataset}, and \algname{} achieves SOTA accuracy, strong controllability, and high scene consistency.

%% file: related_works.tex
\section{Related works}\label{sec:related_works}
\noindent\textbf{Data-driven traffic models.}
Traffic models are typically trained in a supervised learning fashion with driving logs. Most common architectures use an encoder-decoder setup where the encoder takes the agents' history and scene information and outputs a latent distribution. A plethora of architectures have been proposed for traffic models, from the early RNN-based \cite{salzmannIvanovicEtAl2020,AlahiGoelEtAl2016} models, to rasterized models using CNNs \cite{KamenevWangEtAl2022,CuiRadosavljevicEtAl2019}, to GNN-based models \cite{CasasGulinoEtAl2020b,salzmannIvanovicEtAl2020}, and recently transformer-based models have been gaining popularity rapidly\cite{vaswaniShazeerEtAl2017,NgiamCaineEtAl2021scene,SuiZhouEtAl2021,GuSunEtAl2021}. Due to the spatial-temporal nature of trajectory prediction, the transformer architecture needs to perform multi-axes attention. While several works \cite{YuanWengEtAl2021,seffCeraEtAl2023} demonstrated the benefits of using a ``flattened'' attention that allows attention between every two data blocks across multiple axes, it does not scale well when the number of axes increases beyond 2. The more common approach is to use separate attention for each axis \cite{NgiamCaineEtAl2021scene,YuMaEtAl2020,nayakantiAlRfouEtAl2023}, which we refer to as factorized attention. 
Depending on whether the prediction is per-agent or on the joint motion of the whole scene, traffic models can be categorized as node-centric or scene-centric. In general, scene-centric models are more compatible with the downstream planning task \cite{ChenKarkusEtAl2023}, yet more difficult to train due to the extra complexity of modeling interactions between agents. Node-centric models tend to have better coverage of the agent motion (reflected in lower minFDE) \cite{luoParkEtAl2023}, yet such coverage is not beneficial to downstream planning as the joint prediction modes are not defined. Specifically, the ``fake'' joint prediction generated by simply taking the product of marginal node-centric predictions often suffers from poor scene consistency and hurts downstream planning performance.

\noindent\textbf{Interpretable intermediate prediction.}
 As mentioned in the introduction, when the latent space is not interpretable, no supervision signal is available for the latent variable, and training by imitation alone can lead to mode collapse. Several interpretable intermediate signals were proposed to give the model more structure. Goal-conditioning \cite{GuSunEtAl2021,XuChenEtAl2022,ZhaoGaoEtAl2020} has been shown to significantly improve the prediction accuracy and provides stability to a traffic simulation under learned traffic models \cite{XuChenEtAl2022,iglkimEtAl2022}. MTR \cite{shi2022mtr} uses a set of learnable anchors as queries for decoding, which plays a similar role as goal-conditioning. However, the selection of goal prediction resolution can be tricky and these models are typically node-centric due to the difficulty of modeling the joint goal distribution of multiple agents. Intentions (e.g., drive straight, left/right lane change) were used as an intermediate signal in \cite{TangSalakhutdinov2019,zhaofangetal2020,casasLuoEtAl2018}, yet they suffer from ambiguity in their definition. Multipath models \cite{ChaiSappEtAl2019,VaradarajanHefnyEtAl2022} use a fixed set of anchors that can be pre-trained or jointly learned, yet they are not interpretable.

Another popular choice of intermediate feature is to use road lanes. Since vehicles generally follow lanes, lane information such as centerlines \cite{GaoSunEtAl2020}, boundaries and the lane graph topology are important clues to the future motion of the agents. Not only are lanes widely used in the encoding phase via GNNs \cite{LiangYangEtAl2020,salzmannIvanovicEtAl2020} and transformers \cite{nayakantiAlRfouEtAl2023,JiaWuEtAl2023}, they have also been used as an intermediate step towards the final trajectory prediction \cite{CuiCasasEtAl2023,wangYeEtAl2022,zhaofangetal2020}.

\noindent\textbf{Modelling interaction between agents.} Agent interactions may not have a significant influence on prediction accuracy, yet it is critical to scene consistency \cite{ChenIvanovicEtAl2022}, which is important for downstream planning/simulation. While it is standard to encode history interaction with the encoder, future interaction between agents can also be modeled in the decoding process, which is particularly important for scene-centric models. However, since such a decoding process typically requires jointly decoding all agents in the scene, and each latent mode corresponds to one decoding process, the computational complexity becomes an issue as the latent mode increases with the number of agents in the scene. A fixed number of modes \cite{NgiamCaineEtAl2021scene,nayakantiAlRfouEtAl2023,YuanWengEtAl2021} makes the mapping from scene modes to each agent's mode not interpretable, and may not be expressive enough for complicated scenes. Alternatively, \cite{ChenIvanovicEtAl2022} breaks the scene into cliques to reduce the mode cardinality, however not without information loss. 

The idea of homotopy and topological invariance has been studied for the purpose of modeling interaction modes between agents. Strictly speaking, a homotopy refers to a group of trajectories sharing a common starting and end point that can be continuously morphed into one another. It first gained interest in planning with static obstacles \cite{bhattacharya2017path,sontges2017computing,bhattacharyaLikhachevEtAl2012,bhattacharya2010search}, then \cite{MavrogiannisBalasubramanianEtAl2022} used a similar idea to describe the relative motion between two pedestrians, and was extended to identifying interaction modes between agents \cite{RohMavrogiannisEtAl2021,mavrogiannisDeCastroEtAl2022}. Due to the 2D nature of road traffic, winding angle (angular distance) is often used to identify homotopy classes, and \cite{ChenVeerEtAl2023} formally defined the open-end homotopy for 2D agents. 

\noindent\textbf{LLMs for AV.} The recent progress of LLMs has prompted attempts to apply LLMs on autonomous driving, including preliminary works that explore LLMs' understanding of the driving task \cite{tang2023domain}, letting LLMs set sub-goals \cite{sha2023languagempc}, performing behavior planning \cite{mao2023gpt,cui2023drivellm,chen2023driving} and multimodal models involving visual, language, and actions \cite{hu2023gaia}. Given the need for an AV stack to run at a fast frame rate and potentially without internet access, it remains an open question regarding a suitable architecture to leverage and integrate LLMs into AV stacks.

%% file: method.tex
\section{\fullalgname{}}\label{sec:method}

The core idea of \algname{} is to use an interpretable set of modes directly identifiable from the GT future rollout as the latent variable, and we denote such modes as Scene Modes (SM). Scene modes are categorical and consist of two parts: agent2lane (a2l) modes $\lm$ and agent2agent (a2a) modes $\homo$. Under a typical setup for traffic model training where the driving log contains the scene static features, agents' history, and future trajectories, \algname{} first identifies the SM from the driving log, referred to as ground truth scene mode (GTSM). The encoder's goal is to correctly predict GTSM from the scene static features and agents' history. The decoder's goal is to reconstruct the agents' future trajectories under any given SM. We argue that these two modes provide a ``skeleton'' of the scene rollout, and from our experiments, they are expressive enough to represent the multimodal behaviors observed in traffic. It should be noted that the definition of Scene Modes is flexible and one can increase/reduce the mode granularity with alternative mode definitions with the ultimate goal of balancing expressibility and sampling complexity. Next, we elaborate on the detailed definition of Scene Modes and key design elements of \algname{}. 

\subsection{Scene Modes}\label{sec:SM}

\noindent\textbf{a2l modes.} We considered two types of a2l modes. Given $N$ agents, $M$ nearby lane segments provided by either HD maps or online perception, at any given time step, a pairwise a2l mode labels each agent-lane pair with one of the following labels: $\mathtt{l}(\mathbf{x}(t),\mathbf{l})\in\mathcal{PL}:=$ [\textit{NOTON}, \textit{ON}, \textit{AHEAD}, \textit{BEHIND}, \textit{LEFTOF}, \textit{RIGHTOF}, \textit{MISALIGN}], where $\mathbf{x}$ is the state trajectory, $\mathbf{l}$ is the lane object. The scene pairwise a2l mode $\lm\in \mathcal{PL}^{N\cdot M}$, that is, the product of the a2l mode of every agent-lane pair. A second possible a2l mode is the unitary a2l mode, which simply labels which lane an agent is one at a given time step. If the agent is not on any of the lanes, a null label is used. The scene unitary a2l mode is then $\lm \in [0,...,M]^N$, where the set $[0,...,M]$ contains the indices of the M lane segments and 0 indicates null. 

The a2l modes are used in two occasions in \algname{}. First, it is used to tokenize the history agent-lane relationship, which is calculated for each history frame and fed to the encoder. Second, we let the encoder predict the a2l mode $\lm$ at the end of the prediction horizon, and it is used as part of the SM. Both pairwise a2l modes and unitary a2l modes can be applied to the two purposes. Upon experimentation, \algname{} employs the pairwise a2l mode in the encoding process, because they provides more information. For the SM, however, the unitary a2l mode is preferred. Its smaller cardinality is beneficial, significantly simplifying the subsequent importance sampling process.

\noindent\textbf{a2l modes} describe the interaction between agents. We adopt the free-end homotopy from \cite{ChenVeerEtAl2023}, which categorizes the relative motion between two agents into 3 modes: [\textit{S}, \textit{CW}, \textit{CCW}] (\textit{static, clockwise, counterclockwise}):  

\begin{equation}\label{eq:homo_def}
    \mathtt{h}:=\left\{ \begin{aligned} 
    CW, & & \Delta \theta(\mathbf{x}_1,\mathbf{x}_2) < -\hat{\theta}\\
    S, & & -\hat{\theta} \leq \Delta \theta(\mathbf{x}_1,\mathbf{x}_2) < \hat{\theta}\\
    CCW, & & \Delta \theta(\mathbf{x}_1,\mathbf{x}_2) > \hat{\theta}
    \end{aligned} \right.
\end{equation}
where $\mathbf{x}_1$ and $\mathbf{x}_2$ are the trajectories of the two agents under a fixed time window, $\hat{\theta}$ is a fixed threshold, $\Delta \theta$ is the angular distance between the two agents, calculated as
\begin{equation*}
    \Delta \theta(\mathbf{x}_1,\mathbf{x}_2) := \sum_{i=1}^{T-1} \arctan \frac{Y_1^{i+1}-Y_2^{i+1}}{X_1^{i+1}-X_2^{i+1}}-\arctan \frac{Y_1^i-Y_1^i}{X_1^i-X_1^i} .
\end{equation*}
Note that $\mathtt{h}$ is defined over a time period and cannot be computed per frame.
\begin{rem}
There exists symmetry in free-end homotopy, i.e., $\mathtt{h}(\mathbf{x}_1,\mathbf{x}_2) \equiv \mathtt{h}(\mathbf{x}_2,\mathbf{x}_1)$, $\mathtt{h}(\mathbf{x}_1,\mathbf{x}_1)\equiv S$, we use masking in the architecture to enforce symmetry.
\end{rem}
Considering the symmetry,, the scene a2a mode $\homo \in \mathtt{h}^{N \cdot (N-1)/2}$ is the product of the a2a modes of all agent pairs excluding self pairs.

% \begin{figure}
%     \centering
%     \includegraphics[width=0.8\columnwidth]{SceneMode.png}
%     \caption{Illustration of Scene Modes, consisting of a2l mode that designates a lane segment for each agent (can be null when the agent ends up not on any lane segment), and a2a modes that desginates the homotopy class for each pair of agents. The colorization indicates examples of predicted a2l mode probability distribution; the solid and dashed arrows indicates two possible homotopy classes (solid for CW and dashed for CCW), with semantic meaning of the red car yielding/ not yielding to the brown car.}
%     \label{fig:SceneMode}
% \end{figure}

\begin{figure*}
    \centering
    \includegraphics[width=0.7\textwidth]{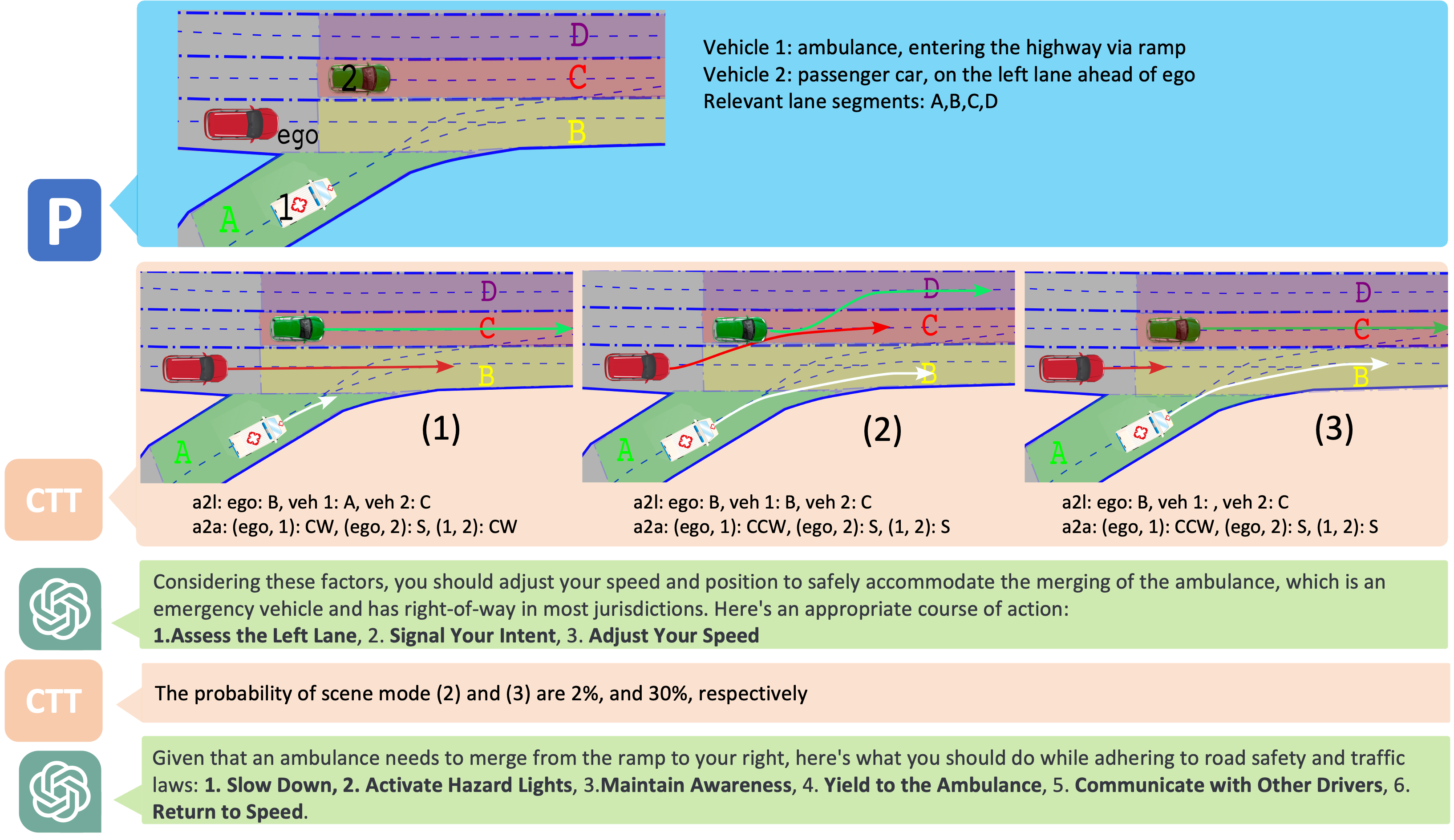}
    \caption{Illustration of integrating \algname{} with GPT-4. The process starts with the perception module providing the scene description, including the road geometry, relevant agents, and relevant lane segments. \algname{} then provides several candidate scene modes, which specify the agent2lane (a2l) and agent2agent (a2a) modes. In this particular scene, the red ego vehicle may choose to yield the ambulance (CCW) or not yield (CW); potential lane changes are specified by a2l modes. Then we query GPT for suggestions, and GPT initially suggests a left lane change to yield to the ambulance while moving forward (SM (2)). However, after checking with \algname{}, the probability of SM (2) is very low, suggesting that human drivers tend to avoid a left lane change under the situation. Eventually GPT takes the feedback from \algname{} and suggests the ego to slow down and maintain lane, i.e., SM (3). The common-sense reasoning of GPT helped eliminate SM(1) whereas the ``expert'' driving knowledge of \algname{} helped eliminate SM (2)}
    \label{fig:LLM_example}
    \vspace{-0.4cm}
\end{figure*}

Scene modes serve as the skeleton of a scene's future evolution and can be directly used to communicate with an LLM. Fig.~\ref{fig:LLM_example} shows an example of integrating \algname{} with an LLM (GPT-4). GPT recognizes that an ambulance is merging in, and that the AV should attempt to yield to the ambulance. However, its initial suggestion of a left lane change was dissuaded by \algname{} as it was deemed dangerous and unhumanlike. With \algname{}'s feedback, GPT modifies its suggestion to ``slow down and maintain lane'', which is more reasonable given the situation. While it may be difficult for an LLM to understand geometry and figure out whether the gap is safe for a lane change, a traffic model like \algname{} is particularly strong with such tasks given that it is trained with traffic data and the encoder is designed to predict scene modes. Since scene modes can be easily parsed into natural language that describes the scene evolution, \algname{} is able to communicate with LLMs. The complete query and answers from GPT can be found in the supplementary material.

Both a2l and a2a modes can be explicitly identified given future trajectories of the agents. In addition, we also calculate margins for each mode, denoted as $\mathcal{M}_\mathtt{l}$ and $\mathcal{M}_\mathtt{h}$. $\mathcal{M}_\mathtt{l}\in\mathbb{R}$ for unitary a2l modes, $\mathcal{M}_\mathtt{l}(\mathbf{x}_i,\mathbf{l}_j)>0$ indicates that agent i is on lane j. $\mathcal{M}_\mathtt{h}(\mathbf{x}_i,\mathbf{x}_j)\in\mathbb{R}^3$, and except for the case where the angular distance is exactly at the threshold, $\mathcal{M}_\mathtt{l}(\mathbf{x}_i,\mathbf{x}_j)$ has exactly one positive entry and two negative entries. The positive entry corresponds to the a2a mode. Both $\mathcal{M}_\mathtt{l}$ and $\mathcal{M}_\mathtt{h}$ are differentiable functions of $\mathbf{x}$ and $\mathbf{l}$ and are used for the consistency loss.

\subsection{Training process}

The encoder of \algname{} encodes the static scene features and agent histories and generates three predictions: the marginal log probability over a2l modes $p_{\lm}$, the marginal log probability over a2a modes $p_{\homo}$, and the joint distribution over the joint SM. When using unitary a2l modes, $p_{\lm}\in \mathbb{R}^{M \cdot N}$, $p_{\homo}\in \mathbb{R}^{3 N \cdot (N-1)/2}$. Since the GT a2l mode and a2a mode are available, $p_{\lm}$ and $p_{\homo}$ can be directly trained as classification tasks with the cross-entropy loss. The joint SM distribution is challenging as the cardinality $|SM|$ scales doubly exponentially with the number of agents: $|SM| = M^N\times 3^{N\cdot (N-1)/2}$, making it computationally infeasible to directly output the whole distribution. 

\noindent\textbf{Importance sampling.} Rather than directly outputting the whole probability distribution over joint SMs, we take an energy function-style approach. Specifically, we train a function that takes in an SM and outputs the unnormalized log-likelihood. This is done on a set of SM samples, and the training objective is to maximize the likelihood of the GTSM after normalization. To generate the set of SM samples, importance sampling is applied. Notice that a joint SM can be decomposed into factors, $N$ a2l factors (using unitary a2l modes) and $N\cdot (N-1)/2$ a2a factors to be precise. In addition, the marginal distributions of these factors are being learned during training, the importance sampling process then boils down to selecting the key factors to generate SM samples. \algname{} employs a 2 stage process.

First, \algname{} scores all factors based on (1) distance to the AV, (2) agent-to-lane distance (for a2l factors), and (3) how concentrated the marginal distribution is. The last score prioritizes factors that are not dominated by one mode. We then fix all the unselected factors to their most likely mode, and select the top K joint modes from the tensor product of the marginal likelihood. Note that the joint probability calculated from the marginal distribution is only an approximation for importance sampling. With the predicted unnormalized log-likelihood for these SM samples, the joint mode loss is a cross-entropy loss
\begin{equation*}
    \mathcal{L}_{SM} = -\log\left(\frac{\exp({g(SM_0)})}{\sum_{i=1}^{K} \exp({g(SM_i)})}\right),
\end{equation*}
where $g(\cdot)$ is the energy function that maps an SM sample to the unnormalized log-likelihood, $\{SM\}_{i=1}^K$ is the set of SM samples with $SM_0$ being the GTSM.

Once the SM distribution is calculated from the encoder, the decoder chooses a set of SM samples (which can be different from the SM samples used for joint SM loss) and conditions the trajectory predictions on those SM samples. During training, one of the SM samples is the GTSM, and all others are diverse samples that differ from the GTSM in a2l, a2a modes, or both. The decoding loss includes a reconstruction loss $\mathcal{L}_{recon}$ and two consistency losses. $\mathcal{L}_{recon}$ is only applied to the decoded trajectories under GTSM, and is calculated as the L2 distance to the ground truth future trajectories of the agents. The two consistency losses are on a2l modes and a2a modes. Specifically, 
\begin{equation*}
    \mathcal{L}_{\lm,con} = \sum_{k=1}^K\sum_{i=1}^N\sum_{j=1}^M \relu{-\mathtt{l}^k[i,j] \cdot \mathcal{M}_\mathtt{l}(\hat{\mathbf{x}}^k_i,\mathbf{l}_j)},
\end{equation*}
where $\hat{\mathbf{x}}^k$ is the trajectory prediction from the decoder under SM sample $k$, and $\mathcal{L}_{\lm,con}$ penalizes negative margins for the agent-lane pairs labeled true in the SM samples. Similarly,
\begin{equation*}
    \mathcal{L}_{\homo,con} = \sum_{k=1}^K\sum_{i=1}^N\sum_{j=1}^N \relu{-\langle\mathtt{h}^k[i,j] , \mathcal{M}_\mathtt{h}(\hat{\mathbf{x}}^k_i,\hat{\mathbf{x}}^k_j)\rangle},
\end{equation*}
which penalizes the negative margins for the selected homotopy classes. The total training loss is then
\begin{equation}\label{eq:loss}
\mathcal{L}=\mathcal{L}_{\lm}+\mathcal{L}_{\homo}+\mathcal{L}_{SM}+\mathcal{L}_{recon}+\mathcal{L}_{\lm,con}+\mathcal{L}_{\homo,con}+\mathcal{L}_{reg},
\end{equation}
where the first two terms are classification losses for the two marginal distributions, $\mathcal{L}_{SM}$ is the joint SM mode classification loss, $\mathcal{L}_{recon}$ is the reconstruction loss under GTSM, followed by two consistency losses. $\mathcal{L}_{reg}$ contains some typical regularization terms such as the L2 regularization, collision losses, and control input regularization.

\subsection{Architecture}
\noindent\textbf{\algname{} Encoder} is based on transformers and GNNs. Since the scene features involve multiple variables with multiple axes, e.g. the temporal axis, agent axis, and lane axis, we designed a standard message passing API that can perform message passing (via either attention or GNN) for any combination of variables on any shared axis. Specifically, there are two types of variables: node variables, including agent history, agent future, and lane segments, and edge variables, including a2l edges and a2a edges. All variables are embedded from raw features with the same embedding dimension $d_e$, and the embedding dimension does not change with message passing or attention updates. 

\begin{rem}
    The reason we add GNNs on top of the transformer encoder is due to the need to explicitly model edges for a2a and a2l mode prediction. We tried implicitly modeling edges with a transformer encoder alone, but the SM prediction performance was poor.
\end{rem}

\begin{figure}
    \centering
    \includegraphics[width=1\columnwidth]{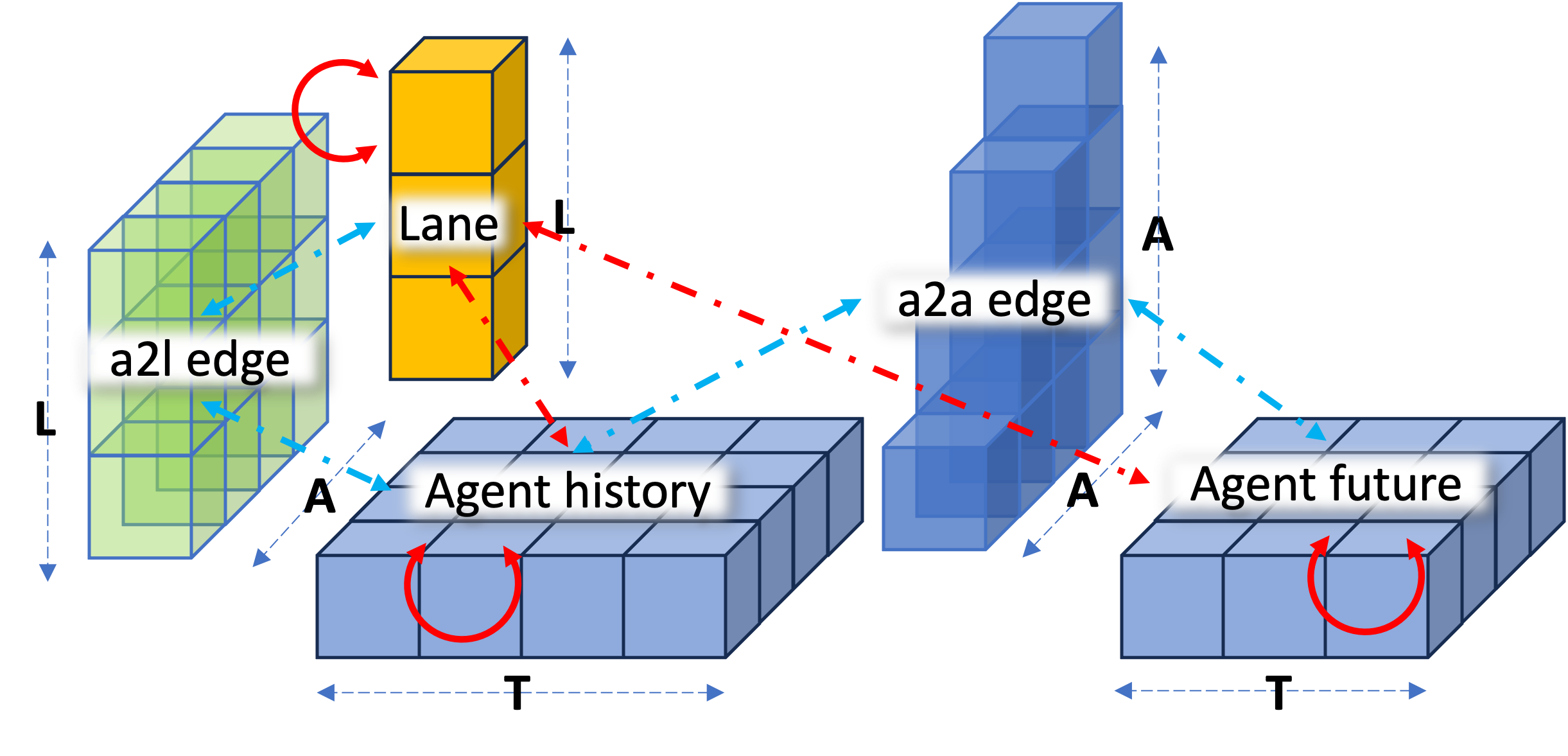}
    \caption{Node variables (solid) and edge variables (transparent) with their axes (\textbf{T}: temporal, \textbf{A}: agent, and \textbf{L}: lane). GNN message passing (cyan dashed arrow), cross-attention (dashed red arrows) and self-attention (solid red arrows) are intertwined.}
    \label{fig:variables}
\end{figure}
\noindent\textbf{}

Figure \ref{fig:variables} shows the 5 key variables and the interactions between them, including GNN message passing, self-attention, and cross-attention. The GNN shares node variables with the transformer, and the GNN message passing and attention updates can be scheduled in arbitrary order.

Similar to \cite{liangyangEtAl2020LaneGCN}, the GNN message passing contains two types of updates. An edge update first concatenates the edge embedding with the two connected node embeddings, then passes through an MLP with a residue connection. Meanwhile, a node update concatenates a node embedding with all the connected edge embedding of a certain type, then passes through an MLP, a pooling layer, and a residue connection. In our experiments we use multi-head attention as the pooling layer with a learnable query token.

A crucial feature to the robustness of traffic models is equivariance, i.e., invariance under coordinate transformation. We extended the relative positional embedding idea \cite{wu2021rethinking} to allow for custom edge functions, and term it Custom Edge Embedding (CEE), which is more flexible and can cater to different variable types. Specifically, all node embeddings are generated with purely local information without any global coordinates. For agents, we concatenate agent types, sizes, and non-coordinate states such as velocity, acceleration, and yaw rate, and embed the raw features to obtain the embedding. Lane segments are given as polylines consisting of multiple waypoints. We use the local coordinate frame centered at the first lane point to represent the shape of the polyline for embedding. While the node embedding is ignorant of its global position, we store the global coordinate in the auxiliary variable for the agent nodes and lane nodes and use a custom edge function with the auxiliary variable as inputs to embed the relative position between the two nodes it connects. Specifically,
\begin{equation}\label{eq:aux_attn}
    \begin{aligned}
        & q=F_q(x),\\
        & k=F_k(\mathbf{Concat}[y,\mathbf{E}_{xy}(x^{\aux},y^{\aux})])\\
        & v=F_v(\mathbf{Concat}[y,\mathbf{E}_{xy}(x^{\aux},y^{\aux})])\\
        & \mathbf{Attn}(x,y) = \text{Softmax}(q^\intercal k \slash\sqrt{d_k})v,
    \end{aligned}
\end{equation}
where $F_{q,k,v}$ are MLPs, $x^{\aux}$ is the auxiliary variable associated with $x$, which can be any node variables, $\mathbf{E}_{xy}$ is the custom edge function for the edge type $xy$. We define custom edge functions for each type of edge based on our insights on what information is important. For example, the a2a edge function calculates the relative position in the local coordinate frame, the a2l edge function output contains the projection of the agent's position on the lane polyline, as well as the starting and ending points of the lane segment in the agent coordinate frame.
\begin{rem}
    We experimented with an attention mechanism among the lane points to obtain the a2l features, and the performance is worse than the projection method.
\end{rem}

The edge function only gets called during the attention calculation and the auxiliary variables do not enter the node embedding. The attention over the time axis is relatively easy as no coordinate frame is involved, we simply use a learnable embedding as positional embedding.

\noindent\textbf{Mode prediction heads} of \algname{} consist of the a2l and a2a mode prediction heads and the joint SM prediction head. The former two are straightforward thanks to the a2l and a2a edges. Recall that a2l and a2a edges are explicitly encoded, and their dimensionalities are $[N,M,T_h]$ and $[N,N,T_h]$ respectively, where $T_h$ is the number of history frames. We perform a pooling operation to squash the time dimension and directly pass them through two MLPs to obtain the log-likelihood of a2l and a2a edges. Specifically, using the unitary a2l modes, $\log \mathbb{P}(\lm)\in\mathbb{R}^{N\cdot M}$. Due to the symmetry of the a2a feature, we average it with its transpose before passing it through the MLP, hence $\log \mathbb{P}(\homo)\in\mathbb{R}^{N\cdot(N-1)/2}$. 

As mentioned in Section.~\ref{sec:SM}, we use an energy-based function approach to generate unnormalized log-likelihood for a given set of SM samples. We first decompose the selected SM samples into a2l and a2a modes, then embed both via two trainable embeddings and concatenate the SM embedding with the a2l and a2a edges from the encoder where the raw edge features from the encoder are tiled to accommodate multiple SM samples. Then, a designated GNN is called to perform several rounds of message passing before the edges are pooled on all three axes (\textbf{T}, \textbf{A}, and \textbf{L}) to obtain a scalar for each SM sample. These are treated as unnormalized log-likelihoods for the joint SM samples.

\noindent\textbf{\algname{} decoder} uses a similar structure as the encoder, with the complication that the agent future trajectory is unknown when performing attention or GNN updates. We consider two strategies: either using an autoregressive procedure and masking out the unknown future blocks, or decoding the whole trajectory in one shot. Under the one-shot strategy, we first fill the auxiliary variables associated with the agents' future blocks with the current position of the agents, then perform multiple rounds of decoding, updating the auxiliary variables after each round. An ablation study to compare the two decoding strategies is included in Section \ref{sec:ablation}. To inform the decoder of the SM it is under, the conditioned a2l modes and a2a modes are appended to the custom edge that enters the multi-head attention shown in Eq.~\eqref{eq:aux_attn}.

% \begin{figure}
%     \centering
%     \includegraphics[width=1\columnwidth]{GNN.png}
%     \caption{Two types of GNN message passing}
%     \label{fig:GNN}
% \end{figure}

%% file: result.tex
\section{Results}\label{sec:result}
\subsection{Performance metrics}
We focus on three aspects of performance when evaluating \algname{}: accuracy, scene consistency, and controllability. We compare \algname{} with recent scene-centric traffic models as benchmarks. In particular, we were able to generate metrics for AgentFormer \cite{YuanWengEtAl2021} directly from a model trained with its codebase.

\noindent\textbf{Accuracy} is measured by the commonly used metrics of Average/Final Displacement Error (ADE/FDE). Luo et al.~\cite{luoParkEtAl2023} note that predicting agents' joint future trajectories (scene-centric) is more challenging than their marginal distributions (node-centric) due to the need to capture interactions between agents. Crucially, metrics like minFDE misleadingly inflates accuracy in marginal predictions: a marginal prediction of a scene with $N$ agents and $m$ modes is akin to a joint prediction with $N^m$ modes. While node-centric predictions can cover a larger area in the trajectory space, leading to low minFDE, it does not translate to practical utility for the downstream planner, which cannot effectively use these combined multimodal marginal predictions.

%\noindent\textbf{Accuracy} is measured by the commonly used metrics of Average/Final Displacement Error (ADE/FDE). As pointed out in \cite{luoParkEtAl2023}, predicting the joint distribution (scene-centric) of all agents' future trajectories is more challenging than predicting the marginal distributions (node-centric). The former needs to consider the interactions between agents and scene-consistency. More importantly, the commonly used metrics of minADE and minFDE give misleading reflection on accuracy for marginal predictions as a marginal prediction with $m$ modes for a scene with $N$ agents is roughly equivalent to a joint prediction with $N^m$ modes when measured with minADE and minFDE. Node-centric models typically cover more area in the trajectory space with the number of modes it predicts with thanks to the minimum operation between modes. However, the seemingly high accuracy is not beneficial to the downstream planner as the planner cannot effectively utilize the combined multimodal marginal predictions of different agents.

On the other hand, while scene-centric models are more compatible with downstream planners, they sacrifice mode diversity on a per-agent level and thus are ``less accurate'' when evaluated using minADE and minFDE. This phenomenon is even more significant in the case of \algname{} as it is designed to generate semantically different modes. We observed that the decoded trajectory for an agent is only affected by its own lane mode and the homotopy w.r.t. some nearby agents. Given that the importance sampling procedure typically samples scene modes that do not differ much (to maximize likelihood), it is common that the predicted trajectories for all but a few agents remain unchanged in between modes. Such consistency is beneficial to the downstream planner (e.g. less noise for homotopy and tree topology identification), yet it hurts minADE and minFDE. 

Despite this disadvantage (although by design), \algname{} achieves SOTA performance on WOMD, and significantly outperforms the benchmarks on nuScenes, as shown in Table~\ref{tab:womd_accuracy} and Table~\ref{tab:nusc_accuracy}. With a longer horizon (8s on WOMD), \algname{} needs to include a large number of lane segments within the driving range of the agents, which made it more challenging to generate accurate predictions.
\begin{table}[]
\centering
\caption{Accuracy of \algname{} and benchmarks on Waymo Open Dataset (WOMD), averaged at horizons of 3s, 5s, and 8s. ST: Scene Transformer, JFP: Joint Future Prediction, AF: Agentformer}
\label{tab:womd_accuracy}
\begin{tabular}{c|cccc}
& ML ADE & minADE & ML FDE & minFDE \\ \hline
ST\cite{NgiamCaineEtAl2021scene} & -      & 1.72    & -      & 3.98    \\
MTR\cite{shi2022mtr}               & -      & 0.92   & -      & 2.06  \\
JFP\cite{luoParkEtAl2023}               & -      & 0.87    & -      & \textbf{1.96}    \\
AF\cite{YuanWengEtAl2021}                & 3.24   & 2.36    & 7.86   & 5.1     \\
\colorcello \algname{}    & \colorcello \textbf{0.97}  & \colorcello \textbf{0.80}   & \colorcello \textbf{2.67}  & \colorcello 2.08   
\end{tabular}
\end{table}
\begin{table}[]
\centering
\caption{Accuracy performance of \algname{} and benchmarks on nuScenes dataset, prediction horizon is 3 seconds.}
\label{tab:nusc_accuracy}
\begin{tabular}{c|cccc}
      & ML ADE & minADE & ML FDE & minFDE \\ \hline
AF\cite{YuanWengEtAl2021}    & 1.23   & 0.8    & 2.63   & 1.6    \\
ILVM \cite{CasasGulinoEtAl2020b}  & -      & 0.86   & -      & 1.84   \\
ScePT \cite{ChenIvanovicEtAl2022} & -      & -      & 1.63   & 1.36   \\
\colorcello \algname{}\tablefootnote{Metrics are from a model trained on nuPlan, evaluated on nuScenes, see Section~\ref{sec:ablation} for detail}   &\colorcello  \textbf{0.55}   & \colorcello \textbf{0.43}   & \colorcello \textbf{1.37}   & \colorcello \textbf{0.96}  
\end{tabular}
\end{table}

\noindent\textbf{Scene consistency} is measured by the collision rate of the predicted trajectories, which is calculated following the geometric computation detailed in \cite{ChenIvanovicEtAl2022}. 
\begin{table}[]
\centering
\caption{Collision rate of the most likely mode and averaged over all modes.}
\label{tab:collision}
\begin{tabular}{c|cccc}
\textbf{} & \colorcello \algname{}          & AgentFormer & ScePT           & Trajectron++ \\ \hline
ML        & \colorcello \textbf{0.27\%} & 2.20\%      & -               & -            \\
all       & \colorcello 0.29\%          & 2.23\%      & \textbf{0.27\%} & 4.20\%      
\end{tabular}
\end{table}
Table~\ref{tab:collision} shows the collision rate of \algname{} and benchmarks under the most likely mode and across all modes. \algname{}'s prediction achieves a lower collision rate than most benchmarks except ScePT, which focuses on scene consistency, but ScePT's inference time is significantly longer than \algname{}.

\noindent\textbf{Controllability} refers to the ability to generate behaviors conditioned on the scene mode. Table~\ref{tab:mode_metrics} shows scene mode prediction and decoding consistency metrics of \algname{}. On nuScenes, the a2a mode accuracy is higher than on WOMD, which uses a longer horizon. The a2l accuracy seems higher on WOMD, yet we suspect it is due to the introduction of a null lane mode, which is the label used when the agent is not on any of the candidate lane segments at the end of the prediction horizon. Due to the long horizon, agents can travel a long distance, and the chance that a null lane mode is used is higher, which increases the lane mode accuracy.

\begin{table}[b]
\centering
\caption{Mode prediction accuracy and consistency rate of \algname{} on nuScenes and WOMD. We show metrics on the marginal a2l mode prediction, a2a mode prediction, and the joint scene mode (SM) prediction. For each metric, the first number is on nuScenes and the second number is on WOMD. Accuracy is the average classification accuracy calculated from the predicted logits; ML correct rate is the average rate that the most likely mode is correct; and consistent rate is the probability that the decoded trajectory satisfies the scene mode it is conditioned on.}
\label{tab:mode_metrics}
\resizebox{\columnwidth}{!}{
\begin{tabular}{c|ccc}
    & Accuracy          & ML correct rate   & consistent rate   \\ \hline
a2l & 76.08\% / 76.81\% & 78.86\% / 83.56\% & 98.13\% / 98.90\% \\
a2a & 89.85\% / 80.98\% & 91.5\% / 84.88\%  & 92.88\% / 85.99\% \\
SM  & 67.11\% / 65.26\% & 78.28\% / 82.23\% &                  
\end{tabular}
}
\end{table}

\begin{table}[]
\centering
\caption{Categorical accuracy of predicted trajectories. Correct rate is the rate at which the ML trajectory prediction matches the GT mode (a2l, a2a, or both).  Cover rate is the rate at which one of the predicted trajectories matches the GT mode. The two numbers are metrics evaluated on nuScenes and WOMD, respectively}
\label{tab:mode_coverage}
\begin{tabular}{c|cc}
                 & AF \cite{YuanWengEtAl2021}               & \colorcello \algname{}                       \\ \hline
a2l correct rate & 74.09\% / 77.66\% & \colorcello \textbf{85.75\% / 87.31\%} \\
a2l cover rate   & 85.82\% / 84.90\% & \colorcello \textbf{90.38\% / 92.64\%} \\
a2a correct rate & 81.66\% / 42.46\% & \colorcello \textbf{88.78\% / 79.28\%} \\
a2a cover rate   & 92.01\% / 43.88\% & \colorcello \textbf{95.01\% / 89.61\%} \\
SM correct rate  & 59.95\% / 35.52\% & \colorcello \textbf{76.24\% / 70.30\%} \\
SM cover rate    & 77.68\% / 39.37\% & \colorcello \textbf{82.26\% / 76.60\%}
\end{tabular}
\end{table}
Table~\ref{tab:mode_coverage} shows the rate of the decoded trajectory under the same SM as the GT trajectory. \algname{} outperforms the benchmark both in ML correct rate and coverage. Especially with the long prediction horizon on WOMD, the gap is larger.
\begin{figure}
    \centering
    \includegraphics[width=1\columnwidth]{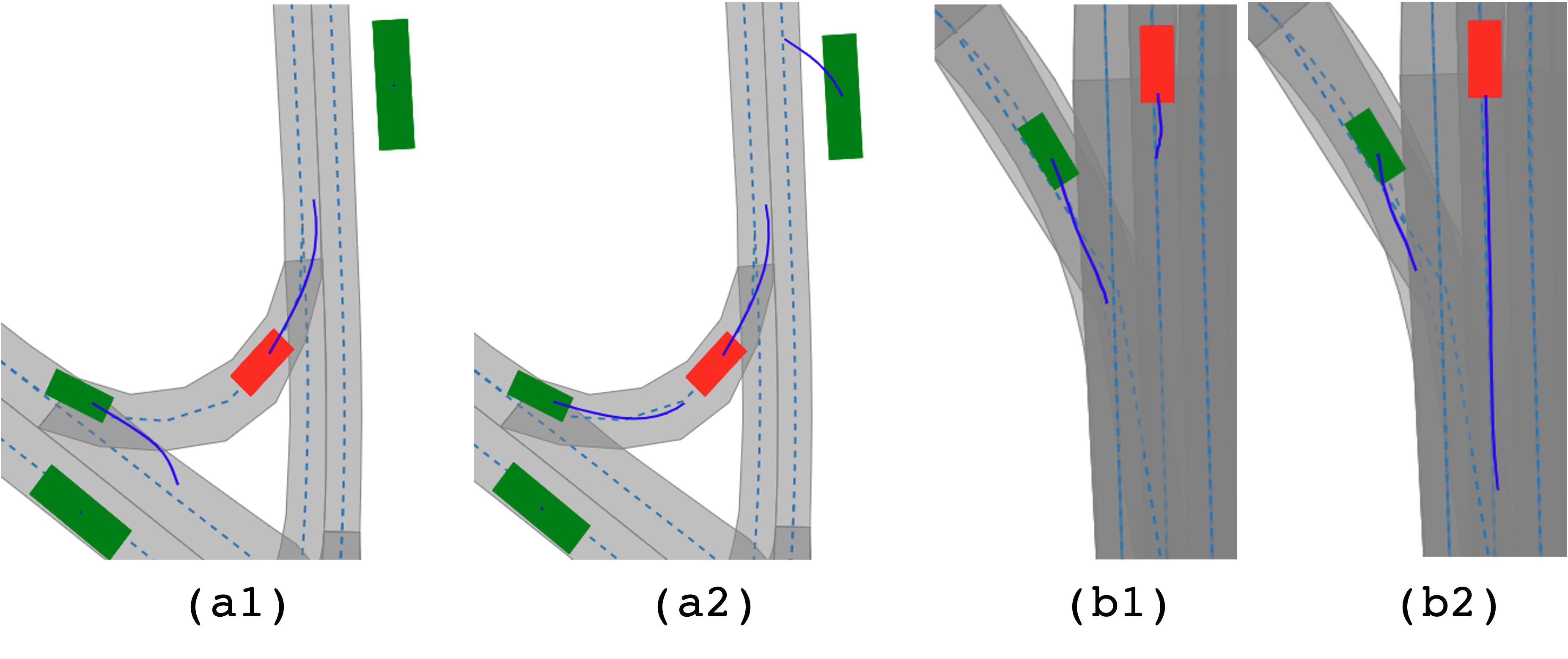}
    \caption{Predictions under different scene modes. The ego vehicle is in red and all other agents are in green. (a1) and (a2) show the impact of lane modes and (b1) and (b2) show the impact of homotopies, where the two homotopies correspond to the ego yielding or not yielding to the green merging vehicle.}
    \label{fig:SM_example}
\end{figure}
Qualitatively, Fig.~\ref{fig:SM_example} demonstrates the controllability of \algname{}. The decoder is able to generate vastly different predictions conditioned on the lane mode and homotopy conditions, even when some of the modes are not realistic, e.g., the parked car suddenly merging into a lane in (a2).

\subsection{Ablation studies}\label{sec:ablation}
\noindent\textbf{Cross-dataset evaluation.}
We train the same \algname{} model on nuScenes training split, which contains 500 scenes, and the training split of nuPlan, which contains 65706 scenes, and evaluate the model on the validation split of the two datasets (for nuplan, we used mini\_val for efficiency).
\begin{table}[]
\centering
\caption{Cross dataset performance of \algname{} between nuScenes and nuPlan}
\label{tab:cross_dataset}
\begin{tabular}{c|cc|cc}
Evaled on      & \multicolumn{2}{c|}{nuScenes}          & \multicolumn{2}{c}{nuPlan mini}   \\ \hline
Trained on     & nuScenes            & nuPlan          & nuScenes            & nuPlan          \\ \hline
minADE         & 0.52            & \textbf{0.43}   & 0.52            & \textbf{0.34}   \\
minFDE         & 1.16            & \textbf{0.96}   & 1.19            & \textbf{0.77}   \\
a2l accuracy   & 75.4\%          & \textbf{75.9\%} & 63.2\%          & \textbf{67.3\%} \\
a2a accuracy   & 90.0\% & \textbf{90.4\%}          & 90.3\% & \textbf{90.8\%}          \\
SM accuracy    & 64.0\%          & \textbf{71.4\%} & 47.4\%          & \textbf{73.2\%} \\
Collision rate & 0.67\%          & \textbf{0.29\%} & 0.54\%          & \textbf{0.16\%}
\end{tabular}
\end{table}

Table~\ref{tab:cross_dataset} shows the cross-dataset performance between nuScenes and nuPlan. The model trained on nuPlan (for about 2 epochs) significantly outperforms the one trained on nuScenes, even when evaluated on nuScenes. This phenomenon shows that \algname{} scales well with data size, and the learned traffic behavior generalizes across datasets.

\noindent\textbf{Decoding strategy.} 
We compared two decoding strategies, autoregressive (AR) and one-shot (OS). The autoregressive decoder runs the transformer encoder 1 step at a time, masking all blocks beyond the current time step. After each transformer run, the decoder generates the position and heading of the agents of only that time step and updates the current block, including the auxiliary variables. The one-shot decoder, on the other hand, generates the full predicted trajectories in one shot. To improve the accuracy, we let the decoder run several rounds of decoding, and update the blocks and auxiliary variables after each round. We let the decoder run 5 rounds in our experiments.
\begin{table}[h]
\centering
\caption{Performance and runtime of autoregressive (AR) decoder and one-shot (OS) decoder on nuScenes and WOMD with different number of future steps.}
\label{tab:AR_OS}
\resizebox{\columnwidth}{!}{
\begin{tabular}{c|ccc|c}
   & \multicolumn{3}{c|}{nuScenes (12 steps)}                                       & WOMD\tablefootnote{Due to the long prediction horizon of WOMD, training an AR decoder requires too much GPU memory, and we only tested the inference runtime.} (40 steps)   \\ \cline{2-5} 
   & \multicolumn{1}{c|}{runtime} & \multicolumn{1}{c|}{minADE} & minFDE & runtime \\\hline
AR & \multicolumn{1}{c|}{295ms}   & \multicolumn{1}{c|}{0.53}   & 1.19   & 1.75s   \\
OS & \multicolumn{1}{c|}{113ms}   & \multicolumn{1}{c|}{0.52}   & 1.16   & 201ms  
\end{tabular}
}
\end{table}

Table~\ref{tab:AR_OS} shows the performance and runtime of the \algname{} model with AR and OS decoders. While the performance difference is negligible, the runtime with OS strategy is significantly shorter than AR, especially for long prediction horizons as in WOMD. Therefore, all the metrics reported in this paper are under the OS decoding strategy.

\noindent\textbf{Ablation on dynamics and diverse lane sample hack.}
We found two techniques that quantifiably improve the performance of \algname{}. The first one is to use dynamics at the end of the decoder instead of directly outputting trajectories. We use a unicycle model for vehicles and cyclists, and a modified unicycle model for pedestrians with a small velocity bound, a large acceleration bound and no bound on the yaw rate. The second technique is to perturb the importance sampling procedure to encourage more diverse lane samples (DLS). Concretely, during importance sampling, for every agent we take the GT lane mode and look for its left and right neighbors if they exist, and set the probability of the two neighbors the same as the GT lane mode. Note that the hack only affects the importance sampling process and will not impact the mode prediction. The more diverse lane mode samples help the decoder to better learn to generate trajectories given the lane mode.
\begin{table}[h]
\centering
\caption{Ablation on the use of dynamic models and diverse lane sample (DLS) hack. Models are all trained on nuScenes.}
\label{tab:ablation_dyn_lane_hack}
\resizebox{\columnwidth}{!}{
\begin{tabular}{l|cccc}
\multicolumn{1}{c|}{}                       & ML ADE & minADE & ML FDE & minFDE \\ \hline
CTT                                        & \textbf{0.57}   & \textbf{0.47}   & \textbf{1.40}   & \textbf{1.02}   \\
CTT w\textbackslash{}o dynamics            & 1.32   & 1.10   & 1.77   & 1.42   \\
CTT w\textbackslash{}o DLS & 0.62   & 0.5    & 1.51   & 1.13  
\end{tabular}
}
\end{table}
As shown in Table~\ref{tab:ablation_dyn_lane_hack}, both techniques help improve the accuracy of \algname{}.

%% file: conclusion.tex
\section{Conclusion and discussion}\label{sec:conclusion}
We present \algname{}, a scene-centric traffic model with an interpretable latent space. Thanks to the interpretability, \algname{}'s encoder and decoder training are decoupled and thus avoids mode collapse. It also allows \algname{} to communicate with LLMs, providing the expert knowledge of driving lacked by LLMs. \algname{} achieves SOTA performance in prediction accuracy across multiple open datasets, and exhibits strong controllability, i.e., generate trajectory predictions given semantic modes, lacking in most traffic models. For future work, we plan to build a pipeline that allows efficient integration of LLMs with traffic models for language-guided simulation and behavior planning.

%% file: appendix.tex
\appendix

\subsection{Custom edge functions}
\algname{} use $x=[X,Y,\sin(\theta),\cos(\theta)]$ to represent an agent position or a lane point where $\theta$ is the heading angle of the agent or the centerline orientation of the lane point. The benefit of using $\sin$ and $\cos$ over raw angle $\theta$ is that the coordinate is irrelevant of the coordinate frame selection as no angle wrapping is performed. The agent auxiliary feature $\aux^a=[x,\mathfrak{s}]$ contains the global coordinate $x$ as well as static features $\mathfrak{s}$ such as agent type, velocity, length and width. The auxiliary feature for a lane segment is simply the $L$ lane points concatenated $\aux^l=\mathbf{Concat}[x_0,...,x_L]$. 

The relative position of $x_2$ to $x_1$ is calculated as 
\begin{equation*}
    \Delta x_{2\to 1} = \begin{bmatrix}
        \cos(\theta_1) (X_2-X_1) + \sin(\theta_1) (Y_2-Y_1)\\
        -\sin(\theta_1) (X_2-X_1) + \cos(\theta_1) (Y_2-Y_1)\\
        \cos(\theta_1)\sin(\theta_2)-\sin(\theta_1)\cos(\theta_2)\\
        \cos(\theta_1)\cos(\theta_2) + \sin(\theta_1)\sin(\theta_2)
    \end{bmatrix}
\end{equation*}
For a2a edge, the edge function $\mathbf{E}_{a2a}$ returns the relative position as well as the static features of agent 2: $\mathbf{E}^{a2a}(\aux^a_1,\aux^a_2) = \mathbf{Concat}[\Delta x_{2\to 1}, \mathfrak{s}_2]$. 

For a2l edge, we project the agent position on the polyline, and the edge contains the projection point, the start and ending points of the polyline in the agent's local frame.

For l2l edge, we take the start and end points of both lane segments, and calculate their relative positions to each other, and concatenate all relative positions.

Note that these edge features are flexibly combined with categorical features such as a2l modes, a2a modes and l2l modes via concatenation during both encoding and decoding, where l2l modes encodes categorical relationship between lane segments such as lane 1 is the \textit{next lane/ previous lane/ left adjacent lane/ right adjacent lane} of lane 2.

\subsection{GPT query example}
\noindent\textbf{Model}: GPT-4-1106-preview

\noindent\textbf{Context}: You are a helpful assistant that helps an autonomous driving system to make decisions. You need to abide by traffic rules and respect driving etiquette.

\noindent\textbf{User}

I'm driving on the right most lane of a 3 lane road, there is a passenger car on the left lane, 10 meters ahead of me. There is a ramp on my right, an ambulance is driving on the ramp, ready to merge into my lane. I'm 10 meters from the ramp merging point, driving at 40 miles per hour. What should I do?

\noindent\textbf{GPT driving}

Given the situation:

You are on the rightmost lane.
A passenger car is on the left lane, 10 meters ahead.
An ambulance is ready to merge from the ramp on your right.
You are 10 meters from the ramp merging point, traveling at 40 mph (~64 km/h).
Considering these factors, you should adjust your speed and position to safely accommodate the merging of the ambulance, which is an emergency vehicle and has right-of-way in most jurisdictions. Here's an appropriate course of action:

Assess the Left Lane: Check the left lane for space to change lanes safely. Since the passenger car is ahead of you, you may have room to merge left if the next vehicle in that lane is at a safe distance behind the car.

Signal Your Intent: If it is safe to do so, and you can merge left without cutting off or endangering other vehicles, signal your intention to change lanes and smoothly transition into the left lane. This would give the ambulance room to merge onto the roadway from the ramp.

Adjust Your Speed: If changing lanes isn't possible or safe because of traffic in the left lane, you should adjust your speed to either create a gap behind you for the ambulance to merge into or to get ahead of the merge point so the ambulance can enter behind you, depending on the specific circumstances and the ambulance's speed.

Considering the ambulance is likely joining the main road directly from a ramp, it may need some space to adjust to the road's speed. If it's not moving with urgency (no flashing lights or sirens), treat it as a standard vehicle, though still with a higher priority due to its function.

Cooperate with the Ambulance: If the ambulance has its lights and/or sirens on, it's indicating an emergency, and you must do your best to facilitate its journey. If it's not possible to change lanes or adjust speed accordingly, maintain your lane and slow down safely to allow the ambulance to maneuver around you. Be mindful not to stop abruptly or create a situation that could lead to an accident.

Stay Alert: Throughout this process, you should remain highly alert, watch for the behavior of other vehicles, and be prepared to take further evasive actions if necessary to ensure everyone's safety.

Follow Traffic Rules: All of this should be done while respecting traffic laws and signals. Avoid any illegal maneuvers, such as crossing solid lines or making erratic moves.

The key is to remain calm, be observant, and make thoughtful decisions that prioritize safety and respect the right-of-way of emergency vehicles. Your actions should always be smooth and predictable to avoid confusing other drivers.

\noindent\textbf{User}

The gap between us and the car on the left lane is too small for a lane change

\noindent\textbf{GPT driving}

If the gap between you and the car in the left lane is too small for a lane change, you have limited options. Given that an ambulance needs to merge from the ramp to your right, here's what you should do while adhering to road safety and traffic laws:

Slow Down: Ease off the accelerator to reduce your speed gently. By slowing down, you can create more space behind you, making it easier for the ambulance to merge onto the road. Remember to check your rearview mirror to make sure that slowing down won't cause an issue for any vehicles behind you. If there is a car immediately behind you, they should also see the ambulance and understand your need to slow down.

Activate Hazard Lights: Briefly using your hazard lights to signal your intent to other drivers can be appropriate to indicate that there is a special situation ahead (in this case, an ambulance merging).

Maintain Awareness: Keep an eye on the ambulance's position and also the surrounding traffic. Be prepared to adjust your actions based on the dynamic situation on the road.

Yield to the Ambulance: If the ambulance is using its sirens and lights, it's requesting the right-of-way. All vehicles should try to give way to the ambulance as it is responding to an emergency.

Communicate with Other Drivers: Use your vehicle's turn signals, brake lights, and if necessary, the horn to communicate with the car ahead and any vehicles behind you, making them aware of the situation.

Return to Speed: Once the ambulance has merged onto the road, smoothly accelerate back to the appropriate speed if the road conditions and speed limits allow.

Remember that while accommodating an emergency vehicle, you still must ensure that your actions do not jeopardize the safety of other road users or yourself. Do not brake harshly or make sudden maneuvers that could cause a collision or result in loss of control. Your goal is to facilitate the ambulance's passage while still maintaining the overall safety and flow of traffic.